\documentclass[numsec,webpdf,modern,large]{oup-authoring-template}%
\usepackage{url}
\usepackage{amsmath}
\usepackage{graphicx}
\usepackage{subcaption}
\usepackage{pifont}
\usepackage{mwe}

\onecolumn 

\graphicspath{{Figures/}}

\theoremstyle{thmstyleone}%
\theoremstyle{thmstyletwo}%
\theoremstyle{thmstylethree}%

\begin{document}

\journaltitle{Bioinformatics}
\DOI{DOI HERE}
\copyrightyear{2024}
\pubyear{2024}
\access{Advance Access Publication Date: Day Month Year}
\appnotes{Paper}

\firstpage{1}


\title[E-ABIN: an Explainable framework for Anomaly detection in BIological Networks]{E-ABIN: an Explainable module for Anomaly detection in BIological Networks}

\author[1, 2]{Ugo Lomoio}
\author[3]{Tommaso Mazza}
\author[4]{Pierangelo Veltri}
\author[1,$\ast$]{Pietro Hiram Guzzi}

\authormark{Lomoio et al.}

\address[1]{\orgdiv{Department of Surgical and Medical Sciences}, \orgname{Magna Graecia University}, \orgaddress{\street{Viale Europa}, \postcode{88100}, 
\state{Catanzaro}, \country{Italy}}}
\address[2]{\orgname{Relatech}, \orgaddress{ \postcode{87036}, \state{Rende}, \country{Italy}}}
\address[3]{\orgdiv{Computational Biology and Bioinformatics Unit}, \orgname{Fondazione Policlinico Universitario A. Gemelli IRCCS}, \orgaddress{\postcode{00168}, \state{Roma}, \country{Italy}}}
\address[4]{\orgdiv{DIMES}, \orgname{University of Calabria}, \orgaddress{\postcode{87036}, \state{Rende}, \country{Italy}}}

\corresp[$\ast$]{Corresponding author: \href{email:email-id.com}{hguzzi@unicz.it}}

\received{Date}{0}{Year}
\revised{Date}{0}{Year}
\accepted{Date}{0}{Year}



\abstract{ The increasing availability of large-scale omics data calls for robust analytical frameworks capable of handling complex gene expression datasets while offering interpretable results. Recent advances in artificial intelligence have enabled the identification of aberrant molecular patterns distinguishing disease states from healthy controls. Coupled with improvements in model interpretability, these tools now support the identification of genes potentially driving disease phenotypes. However, current approaches to gene anomaly detection often remain limited to single datasets and lack accessible graphical interfaces. Here, we introduce E-ABIN, a general-purpose, explainable framework for Anomaly detection in Biological Networks. E-ABIN combines classical machine learning and graph-based deep learning techniques within a unified, user-friendly platform, enabling the detection and interpretation of anomalies from gene expression or methylation-derived networks. By integrating algorithms such as Support Vector Machines, Random Forests, Graph Autoencoders (GAEs), and Graph Adversarial Attributed Networks (GAANs), E-ABIN ensures a high predictive accuracy while maintaining interpretability. We demonstrate the utility of E-ABIN through case studies of bladder cancer and coeliac disease, where it effectively uncovers biologically relevant anomalies and offers insights into disease mechanisms. This software is freely available at: \url{https://github.com/UgoLomoio/E-ABIN}}

\maketitle

\section{Introduction}

The rapid expansion of high-throughput technologies has profoundly reshaped the analysis of biological and medical data, enabling researchers to extract meaningful insights from large-scale gene expression datasets \cite{asnicar2024machine}. These datasets not only facilitate the exploration of complex biological questions but also support the identification of novel biomarkers \cite{rosati2024differential}. A widely adopted strategy involves representing gene–gene interactions as graphs, which can then be analyzed using graph-based anomaly detection techniques to uncover hidden patterns within biological networks \cite{Xiaoxiao2023gad}. Such graph representations provide a powerful framework for studying gene expression dynamics, detecting mutations, and assessing responses to pharmacological interventions \cite{zamanzadeh2024deep}. Within this context, anomaly detection focuses on identifying nodes, edges, or substructures that exhibit behaviors significantly deviating from expected patterns \cite{Akoglu2014GraphBA}. Additionally, graph-based methods are frequently employed in node classification tasks, leveraging structural or topological information to infer node labels \cite{liu2022bond}.

Current graph anomaly detection methodologies are generally categorized into three main approaches: (i) statistical, (ii) machine learning (ML), and (iii) deep learning (DL) techniques. Statistical methods rely on descriptive graph properties—such as node degree or clustering coefficients—to detect outliers and characterize network structure \cite{rousseeuw2018statistical, BANSAL2022724}. ML-based approaches have been successfully applied to both tabular classification tasks (e.g., healthy vs. diseased states) \cite{rambaud2023binaryclass} and network-based classification. Tools such as CSAX \cite{noto2015csax} detect disrupted expression pathways in individual samples and use ML models to systematically identify anomalies within biological networks. However, these approaches often rely on hand-crafted features that require domain expertise for validation.

To overcome these limitations, DL methods such as graph neural networks (GNNs) and autoencoders (AEs) have emerged as scalable solutions for modeling large and complex biological graphs. Architectures like Deep GONet \cite{bourgeais2021deep} and XOmiVAE \cite{withnell2021xomivae} automatically learn high-level representations, achieving strong predictive performance while enabling a degree of interpretability. Explainability has become especially critical in biomedicine, where trust in model outputs directly impacts clinical or experimental decisions. Recent frameworks such as DSAF-GS \cite{morabito2023genes} and XOmiVAE \cite{withnell2021xomivae} illustrate how deep models can be augmented with mechanisms to generate biologically meaningful explanations, bridging the gap between accuracy and interpretability.

Despite these advancements, current tools often lack user-friendly interfaces or integrated explainability features, limiting their accessibility and utility in applied research settings. To address these limitations, we introduce E-ABIN (Explainable Anomaly detection in Biological Networks), a novel AI-driven framework designed to enhance the interpretability and usability of gene network analysis. E-ABIN offers several key innovations: (i) a graphical user interface (GUI) designed for ease of use; (ii) support for DNA methylation data, in addition to gene expression; (iii) an integrated explainability module that aids biological interpretation of results; and (iv) the ability to construct patient-specific networks, enabling personalized analysis \cite{brooks2024challenges}. E-ABIN combines machine learning and deep learning capabilities within an explainable framework, aiming to outperform existing tools for the analysis of network-encoded gene–pathology associations. Furthermore, its flexible architecture allows it to process a variety of array-based data types. Table~\ref{tab:related} summarizes and compares the features of E-ABIN against current state-of-the-art systems, highlighting its distinctive support for GUI accessibility, explainability, ML/DL integration, and methylation data analysis.

\begin{table}[H]
\centering
\caption{Comparison with respect to State-of-the-Art (SOTA) Software}
\label{tab:related}
\begin{tabular}{| l | l | l | l | c | c |}
\hline
Software & GUI & ML & DL & Explainable & DNA Methylation \\
\hline
XOmiVAE \cite{withnell2021xomivae} & \ding{55} & \ding{55} & \ding{51} & \ding{51} & \ding{55} \\
Deep GONet \cite{bourgeais2021deep} & \ding{55} & \ding{51} & \ding{51} & \ding{51} & \ding{55} \\
DSAF-GS \cite{morabito2023genes} & \ding{55} & \ding{55} & \ding{51} & \ding{51}  & \ding{55} \\
GenePattern \cite{kuehn2008using} & \ding{51} & \ding{51} & \ding{55} & \ding{55}  & \ding{55} \\
CSAX \cite{noto2015csax} & \ding{55} & \ding{51} & \ding{55} & \ding{55} & \ding{55} \\
E-ABIN & \ding{51} & \ding{51} & \ding{51} & \ding{51} & \ding{51} \\
\hline
\end{tabular}
\end{table}

From a functional perspective, E-ABIN advances the state of the art by integrating a graphical user interface (GUI), machine learning (ML), deep learning (DL), and explainability into a unified and accessible framework. This integrated design improves the performance of anomaly detection while enabling users—both researchers and clinicians—to interpret the results in a biologically meaningful context. Unlike existing tools, E-ABIN supports multiple analytic strategies, offering flexibility in the choice of models for graph-based anomaly detection and binary classification on gene expression data. Notably, its embedded explainable AI modules provide critical insight into the underlying biological mechanisms associated with detected anomalies, helping to bridge the gap between computational results and clinical or experimental interpretation \cite{ding2022explainability, anguita2020explainable, toussaint2024explainable}.

E-ABIN’s explainability framework operates at multiple interpretive levels. At the patient level, explanations are tailored to individual subjects, offering personalized insights into network behavior. Depending on the model employed, such as convergence-divergence networks or Individual-Specific Networks (ISNs), the resulting explanation is either node-level or graph-level, respectively. In convergence-divergence networks, each node corresponds to a patient, facilitating node-level interpretation, whereas in ISNs, each graph represents a patient, and thus the explanations reflect graph-level structures. At the disease level, explanations are aggregated across patients within the same class (e.g., “diseased” or “healthy”), yielding high-level insights into class-specific patterns. E-ABIN further supports dataset-level explanations that characterize the model’s global behavior across classes, enhancing interpretability and transparency on a broader scale.

Central to E-ABIN’s architecture is its ability to detect anomalies in biological networks by generating expressive graph representations of gene expression profiles. This is achieved through the use of both convergence/divergence models and the Individual-Specific Network (ISN) paradigm \cite{KUIJJER2019226}, which constructs personalized networks that capture gene–gene relationships unique to each patient. These representations facilitate the identification of both inter-patient and intra-patient variations, improving the sensitivity of the system to subtle, complex, and heterogeneous anomalies across different biological states.

\begin{figure}
    \centering
    \includegraphics[width=0.8\linewidth]{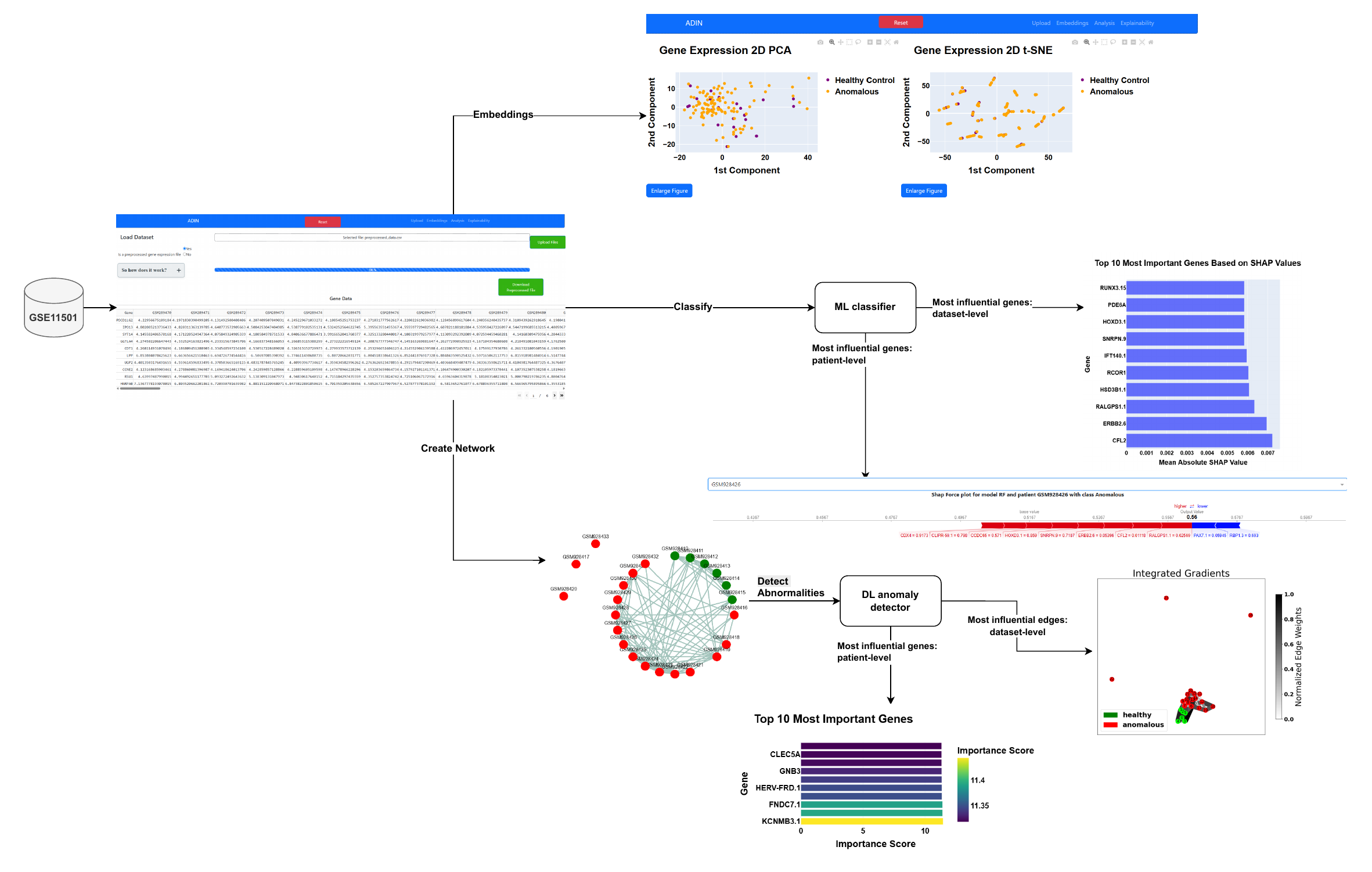}
    \caption{Example of E-ABIN: the user uploads the dataset GSE11501 and performs anomaly detection to detect celiac patients using ML and DL techniques. After that, Explainability methodologies are used to extract the top 10 most influential genes for disease prediction, which will be visualized as a bar plot.}
    \label{fig:usage}
\end{figure}

Figure \ref{fig:usage} illustrates a practical application of E-ABIN in the context of coeliac disease analysis. In this example, the user uploads the publicly available GSE11501 dataset, which contains gene expression profiles from both healthy individuals and patients with coeliac disease. Once loaded, E-ABIN performs anomaly detection and constructs the corresponding association network. As shown in the figure, the tool enables users to identify outlier profiles—such as those indicative of disease—and to interpret these predictions through both patient-level and dataset-level explanations, as also depicted in Fig. \ref{fig:workflow}.

E-ABIN bridges the gap between computational analysis and biological insight by offering an intuitive graphical interface that visualizes the structural and functional anomalies identified in the data. The integrated explainability module, leveraging widely used frameworks such as SHAP \cite{shap} and Captum \cite{kokhlikyan2020captum}, highlights the genes, nodes, or interactions that contribute most strongly to model predictions. This functionality is especially valuable in translational research, where understanding the molecular basis of predictions can support clinical decision-making and guide therapeutic development. As demonstrated in Fig. \ref{fig:usage}, the system reveals key genes implicated in coeliac disease by generating a gradient-based network, distinguishing between disease-associated and non-disease-associated molecular patterns.

We present the architecture and core functionalities of E-ABIN, highlighting its application to gene expression and DNA methylation datasets associated with human pathologies. The tool integrates robust preprocessing pipelines, comprehensive visualization capabilities, and classification modules that streamline the analytical workflow. Advanced machine learning and deep learning models deliver high predictive performance, while built-in interpretability features ensure transparency and facilitate biological interpretation. Through a series of case studies, we demonstrate E-ABIN’s ability to detect meaningful anomalies in biological networks and to provide clear, interpretable explanations.

E-ABIN is designed to be broadly applicable across diverse biomedical contexts. Its flexible architecture allows users to analyze any gene expression dataset presented in tabular form and annotated with class labels, including those from widely used repositories such as GEO. Moreover, the platform supports user-supplied custom datasets, provided they adhere to standard formats (e.g., series matrix files), thus enabling seamless integration into varied research workflows. By minimizing preprocessing requirements and maximizing compatibility, E-ABIN facilitates the adoption of explainable AI approaches in studies of personalized medicine, rare diseases, and systems biology.

\section{E-ABIN Architecture}
\label{sec:architecture}
 
\begin{figure}
    \centering
    \includegraphics[width=\linewidth]{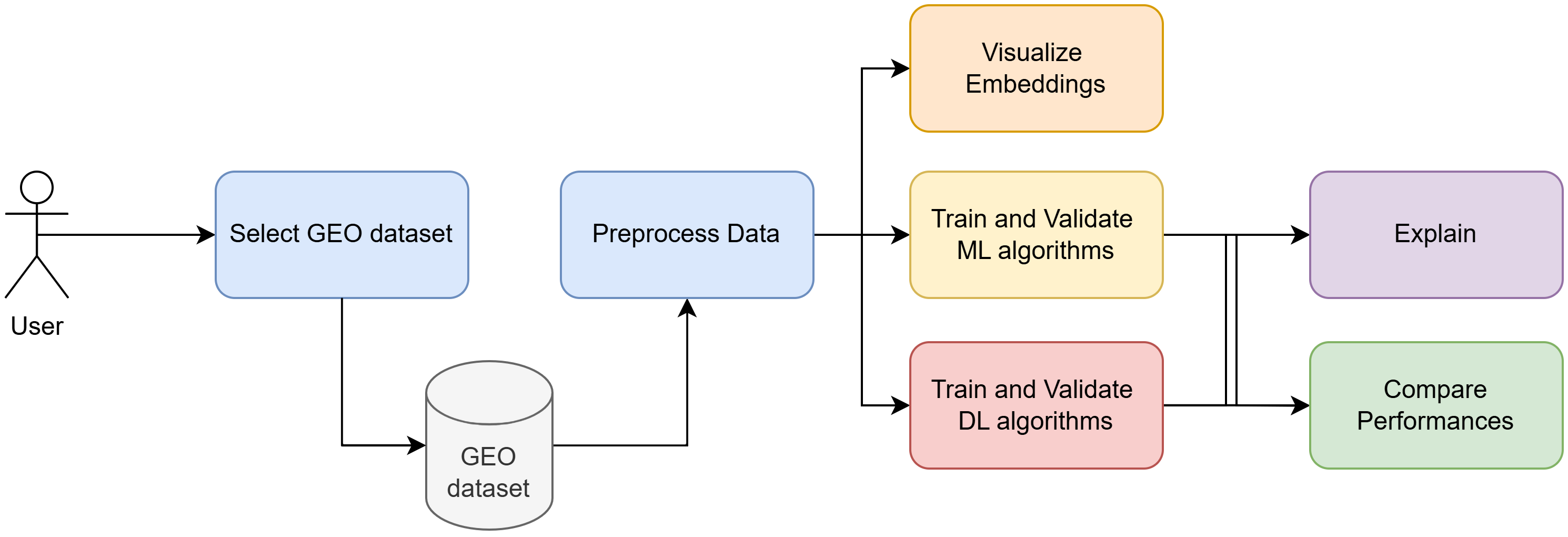}
    \caption{E-ABIN Modules: functionalities and workflow of the E-ABIN available modules}
    \label{fig:workflow}
\end{figure}

Figure~\ref{fig:workflow} presents the overall workflow implemented in \textbf{E-ABIN} for biological data analysis. Starting from a dataset containing gene expression or DNA methylation profiles from both healthy individuals and patients exhibiting anomalous patterns, E-ABIN first applies preprocessing procedures to ensure data quality and consistency. Following this step, the cleaned dataset can be visualized in a reduced two-dimensional space using dimensionality reduction techniques, facilitating initial data exploration and class separability assessment.

Users can then train and evaluate binary classification models to predict whether a given sample corresponds to a healthy profile or to one associated with molecular anomalies. E-ABIN supports both node-level and graph-level deep learning–based anomaly detection by employing multiple graph-representation strategies. In the node-level approach, the entire dataset is encoded into a single graph in which each node represents a patient and the corresponding gene expression or methylation profile is embedded as node features. In the graph-level approach, E-ABIN constructs a separate graph for each patient using the Individual-Specific Network (ISN) paradigm, enabling fine-grained patient-specific analysis and anomaly detection. These dual perspectives offer flexibility in capturing both global patterns and individualized molecular disruptions.

The functional architecture of \textbf{E-ABIN} is designed as a modular framework, with each component contributing to a seamless pipeline that spans from data preprocessing to anomaly detection. This modularity ensures scalability and flexibility, enabling the integration of diverse analytical tasks while maintaining a coherent user experience. The system incorporates a user-friendly Graphical User Interface (GUI), implemented using Python and modern web technologies, allowing users to access the full suite of functionalities without requiring advanced technical expertise.

\begin{figure}[!h]
    \centering
    \includegraphics[width = 0.5\textwidth]{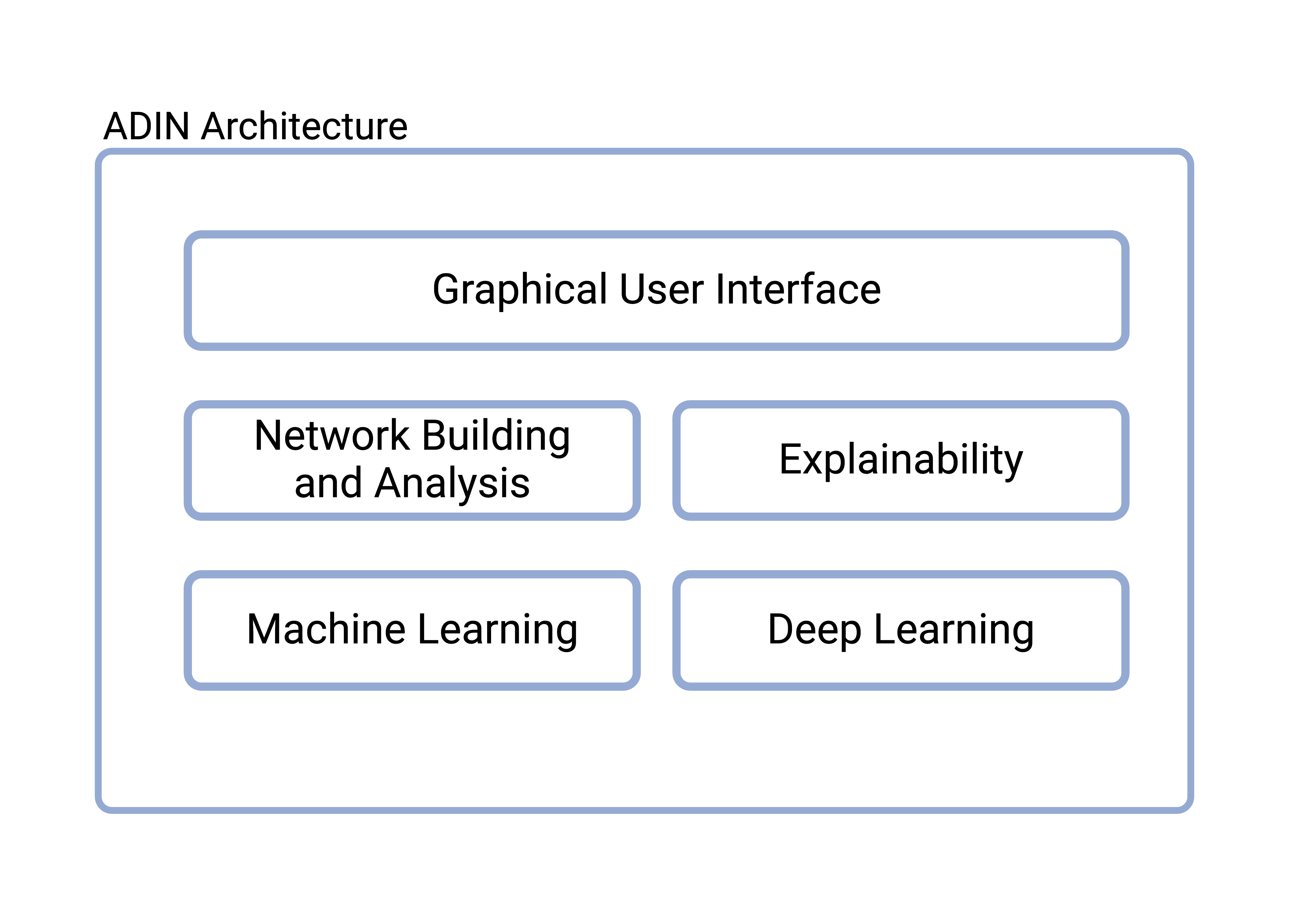}
    \caption{E-ABIN architecture depicting key components: a GUI layer integrated with modules for Network Building and Analysis, Explainability, Machine Learning, and Deep Learning.} 
    \label{fig:architecture}
\end{figure}

Figure~\ref{fig:architecture} illustrates the overall architecture, highlighting the GUI components through which users can access the core modules. These modules collectively form an end-to-end analytical pipeline, capable of handling the complexities of processing and interpreting high-dimensional biological data. By integrating conventional machine learning approaches, graph-based network construction techniques, and state-of-the-art deep learning algorithms, the framework provides a comprehensive and efficient solution for anomaly detection in molecular networks.

\begin{figure}
    \centering
    \includegraphics[width=\linewidth]{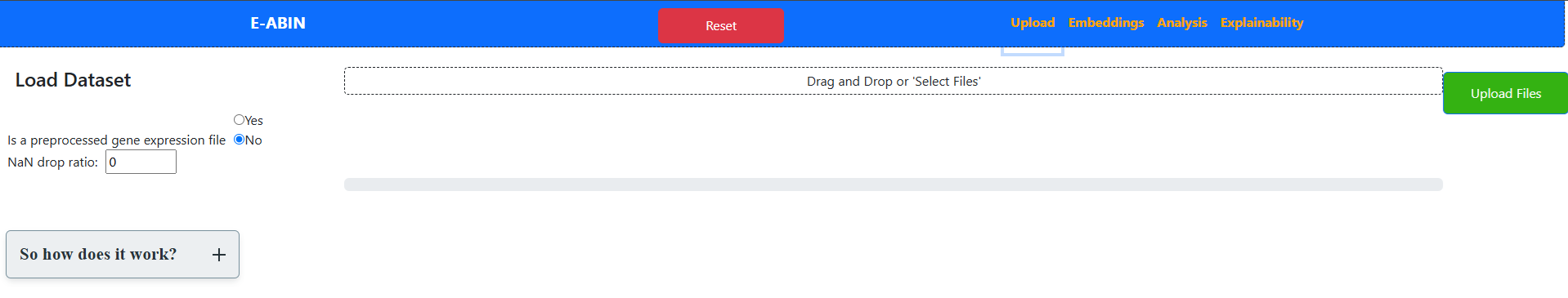}
    \caption{The landing page of E-ABIN GUI. From this page the user can upload a GEO raw dataset or a dataset already preprocessed by E-ABIN. After data uploading, user can navigate between different modules (Upload, Embeddings, Analysis and Explainability) using the navigation bar at the top of the page. At the end of the analysis, the reset button can be used to delete all the current local variables to make the software available for another analysis with a different dataset.}
    \label{fig:gui}
\end{figure}

The GUI, depicted in Figure~\ref{fig:gui}, enables users to upload datasets, configure preprocessing parameters, and launch machine learning or deep learning workflows. Furthermore, it offers real-time visualization of analytical outputs, including network structures, classification outcomes, anomaly scores, and performance metrics. Interactive elements—such as graphs, tables, and charts—facilitate intuitive exploration of results, helping users interpret and communicate their findings effectively. By embedding visualization and analysis capabilities directly within the interface, E-ABIN enhances usability, accelerates discovery, and supports efficient data-driven decision-making across a variety of biomedical contexts.

The \textbf{Network Building and Analysis Module} manages the core data analysis workflow within the \textbf{E-ABIN} framework. Raw datasets—typically sourced from public repositories such as the Gene Expression Omnibus (GEO) database \cite{Clough2016geo}—frequently contain inconsistencies that necessitate preprocessing. This preparatory phase includes the removal of duplicate entries, imputation of missing or null values, and normalization of gene expression or methylation profiles to ensure a uniform scale across features. These steps are essential for enhancing the reliability and interpretability of downstream machine learning (ML) and deep learning (DL) analyses.

A defining innovation of E-ABIN lies in its ability to transform numerical biological data into structured graphs through the construction of \textit{attributed networks}. Leveraging the convergence/divergence network paradigm \cite{zanin2018}, this module encodes similarity and divergence patterns between molecular profiles into a graph-based structure, thereby capturing underlying biological relationships. These graphs serve as input to the DL-based anomaly detection pipeline, enabling the identification of abnormal nodes—representing individual patients or genes—that deviate from expected topological patterns, potentially indicating disease-associated anomalies.

The \textbf{Machine Learning Module} is dedicated to binary classification tasks using gene expression or DNA methylation profiles to distinguish between healthy and diseased samples. The module supports the evaluation of several standard ML algorithms, including logistic regression, decision trees, random forests, k-nearest neighbors, and support vector machines. By benchmarking these models across commonly used performance metrics—such as accuracy, precision, recall, and F1-score—the module ensures rigorous model selection while addressing the challenges posed by the high dimensionality of molecular data.

\begin{figure}[!h]
    \centering
    \includegraphics[width=1.0\linewidth, keepaspectratio]{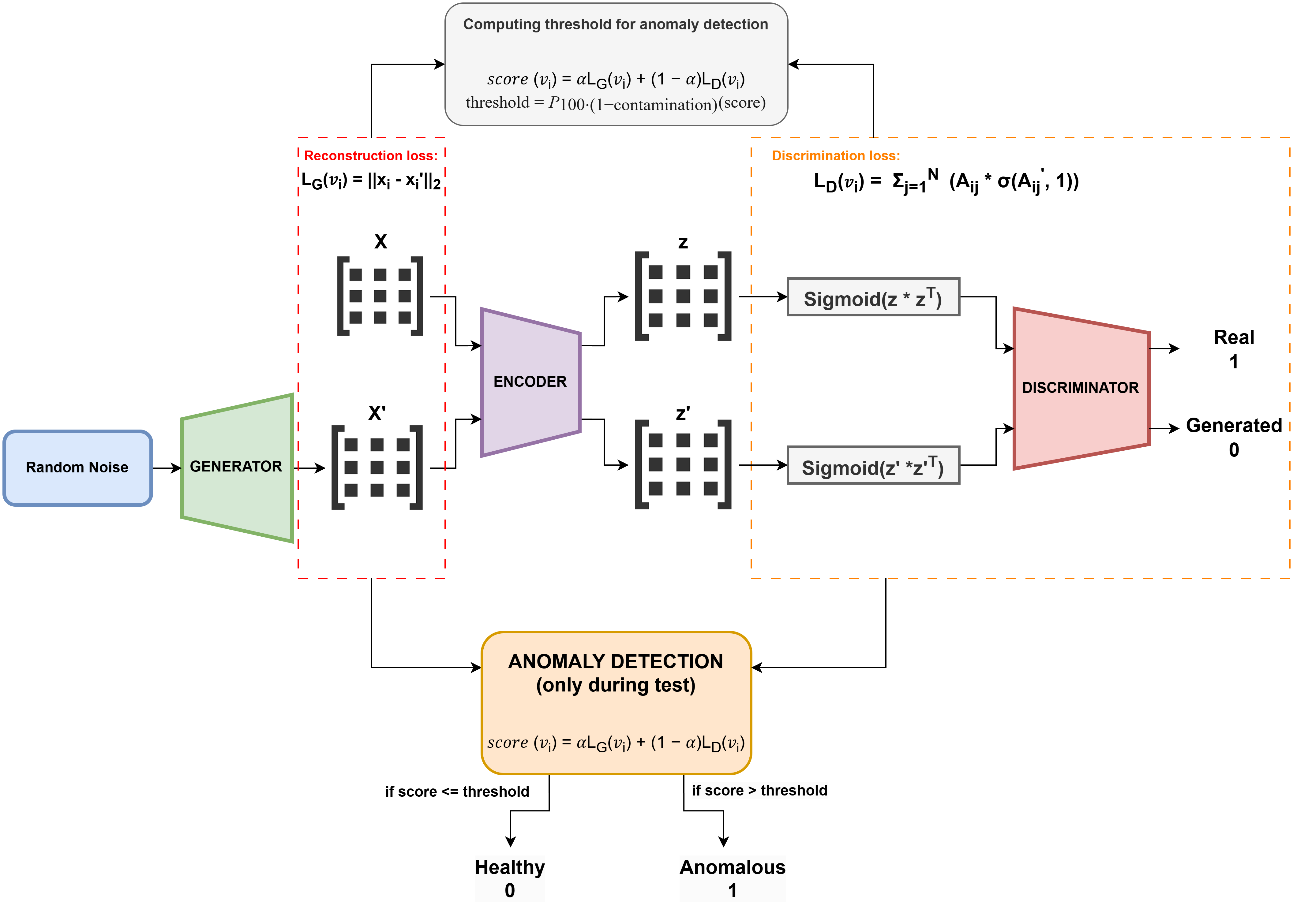}
    \caption{Functional architecture for the Generative Adversarial Attributed Network framework. The Generator samples generated node features x' from a prior Gaussian noise distribution. Then, the Encoder creates low-dimensional embeddings (z and z') for both real and artificial node features (x and x'). The discriminator is finally trained to classify node embeddings into Real or Generated categories. During training, a reconstruction loss $L_G$ is optimized to generate realistic nodes. In contrast, a discriminator loss $L_D$ is used to make a discriminator that can efficiently distinguish between real and generated nodes. Those losses are used to compute the outlier score of each node $v_i$. This score is used during training to compute the threshold value used at test time to perform a threshold-based anomaly detection task. $P_q(score)$ denotes the q-th percentile values in the array score.}
    \label{fig:gaan}
\end{figure}

For advanced network analysis, the \textbf{Deep Learning Module} implements a graph-based anomaly detection strategy using models provided by the \texttt{pygod} library \cite{pygod}. Central to this module is the Generative Adversarial Attributed Network (GAAN), a deep learning architecture specifically designed for detecting anomalies in attributed graphs. GAAN performance is systematically compared with alternative graph-based deep models, including Graph AutoEncoders (GAE) \cite{gae} and Graph Convolutional Networks (GCN) \cite{gcn}. The architecture of the GAAN model, detailing its adversarial structure and feature learning process, is illustrated in Figure~\ref{fig:gaan}.

The GAAN and GAE models were first trained on \textit{Healthy} and \textit{Anomalous} nodes to learn the underlying distribution and patterns of normality of the data. Once trained, the model evaluates the new nodes in the graph by comparing their behaviors to the learned baseline patterns. Nodes that exhibit significant deviations from these patterns are flagged as "Anomalous." This module also compares the performances obtained using GNNs against those obtained with traditional ML, illustrating the benefits of adopting graph-based deep learning for complex anomaly detection tasks.

An important parameter that must be properly configured in GAAN and GAE models is the \textit{contamination} of the dataset, simply computed as the proportion of outliers in the dataset:

\begin{equation}
    \text{contamination} = \frac{N_1}{N_0 + N_1} \times 0.5
    \label{eq:contamination}
\end{equation}

where:
\begin{itemize}
    \item \( N_0 \) is the count of samples in the training set belonging to class \( 0 \).
    \item \( N_1 \) is the count of samples in the training set belonging to class \( 1 \).
\end{itemize}

Finally, the explainability module was used to interpret the model predictions. Predictions obtained by both ML and DL modules are thus explained supporting the comprehension of the results of biological and clinical-related disease studies. This allowed us to better understand and enhance trustworthiness. Shap is used to interpret predictions obtained by ML algorithms with the aim of extracting key genes that play an important role in the studied disease or for a given patient. The predictions obtained using graph-based DL methods can also be explained by using explainability techniques such as GNNExplainer, Integrated Gradients and Saliency Maps. With GNNExplainer we can predict, like Shap, key genes with important role in disease progression. Explainability techniques such as Integrated Gradients and Saliency Maps are available to highlight relations between genes that are important for disease prediction. 

\section{Experimental results}
\label{sec:casestudies}

We tested the E-ABIN framework on a cancer dataset to demonstrate the ability of the analysis in terms of   anomaly detection in both gene expression and methylation networks. 
We consider node-level anomaly detection using GNNs and convergence divergence networks. We also performed graph-level anomaly detection using GNNs and interactome-based individual-specific networks. For both datasets, additional information regarding the graphs obtained can be found in Table \ref{tab:graphs}.

\begin{table}[]
    \centering
    \begin{tabular}{|c|c|c|c|c|c|}
        \hline
        Dataset & Network Type & N. of networks & Nodes & Edges & Attributes \\ \hline
        GSE37817 & convergence-divergence & 1 & 24 & 193 (threshold of 0.93) & 27552 \\ \hline
        GSE37817 & ISNs & 24 & 27552 & 34918 (threshold of 0.98) & 1 \\ \hline
          
    \end{tabular}
    \caption{Additional information for the obtained networks.}
    \label{tab:graphs}
\end{table}

The classification performances obtained in both subtasks were compared with those obtained using baseline machine learning binary classifiers.
For the bladder cancer detection network, a small dataset was selected from the GEO data repository. The selected dataset with accession code GSE37817 contained methylation profiles by array obtained using the GPL8490 Illumina Methylation27 beadchip. Profiling was performed to differentiate between normal cells and from Non-Muscle-Invasive Bladder Cancer (NMIBC). The dataset contains 27552 values for DNA methylation for 24 patients only. Given the limited number of patients, we expect machine-learning models to outperform deep-learning models \cite{pygodbench}.
Table \ref{tab:dataset} lists the unbalanced diagnostic class distributions inside the dataset. 

\begin{table}[h]
    \centering
    \small
    \begin{tabular}{|c|c|}
    \hline
         \textbf{Label} &  \textbf{Count} \\ \hline
         Normal patient & 6 \\ \hline
         Tumour NMIBC patient & 18 \\ \hline
    \end{tabular}
    \caption{GSE37817 dataset label distribution.}
    \label{tab:dataset}
\end{table}

\begin{figure*}[!h]
        \centering
        \begin{subfigure}[b]{0.475\textwidth}
            \centering
            \includegraphics[width=0.7\textwidth]{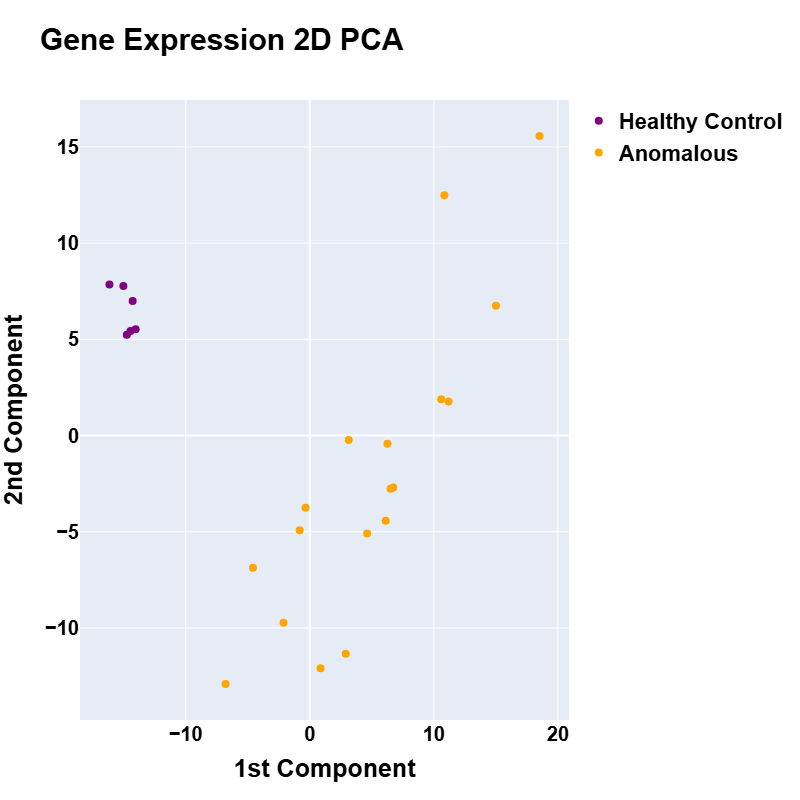}
            \caption[A]%
            {{\small PCA embeddings}}    
            \label{fig:pca_bc}
        \end{subfigure}
        \hfill
        \begin{subfigure}[b]{0.475\textwidth}  
            \centering 
            \includegraphics[width=0.7\textwidth]{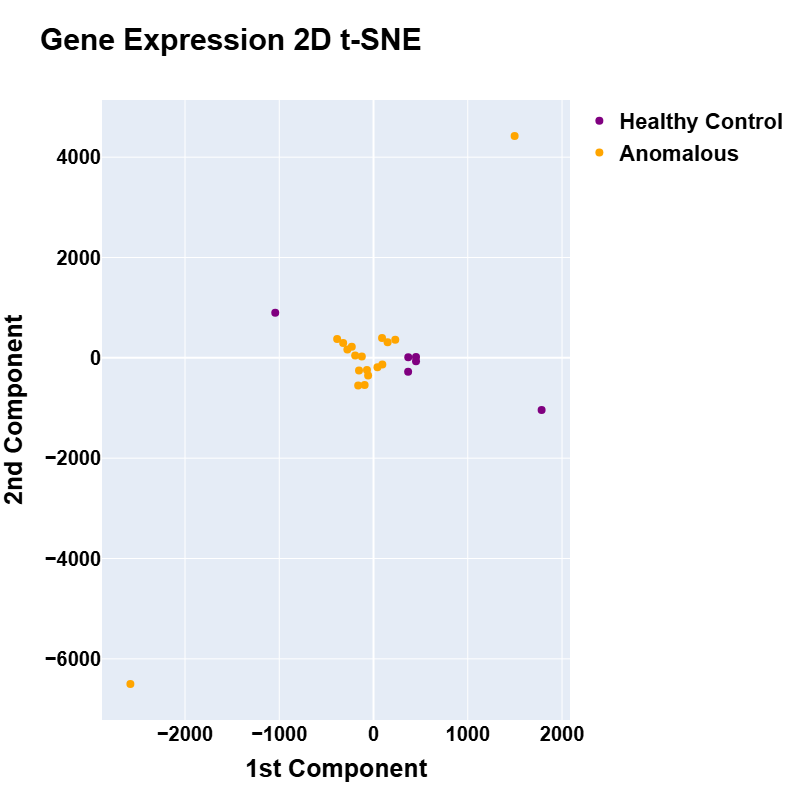}
            \caption[B]%
            {{\small TSNE embeddings}}    
            \label{fig:tsne_bc}
        \end{subfigure}
        \vskip\baselineskip
        \begin{subfigure}[b]{0.475\textwidth}   
            \centering 
            \includegraphics[width=0.7\textwidth]{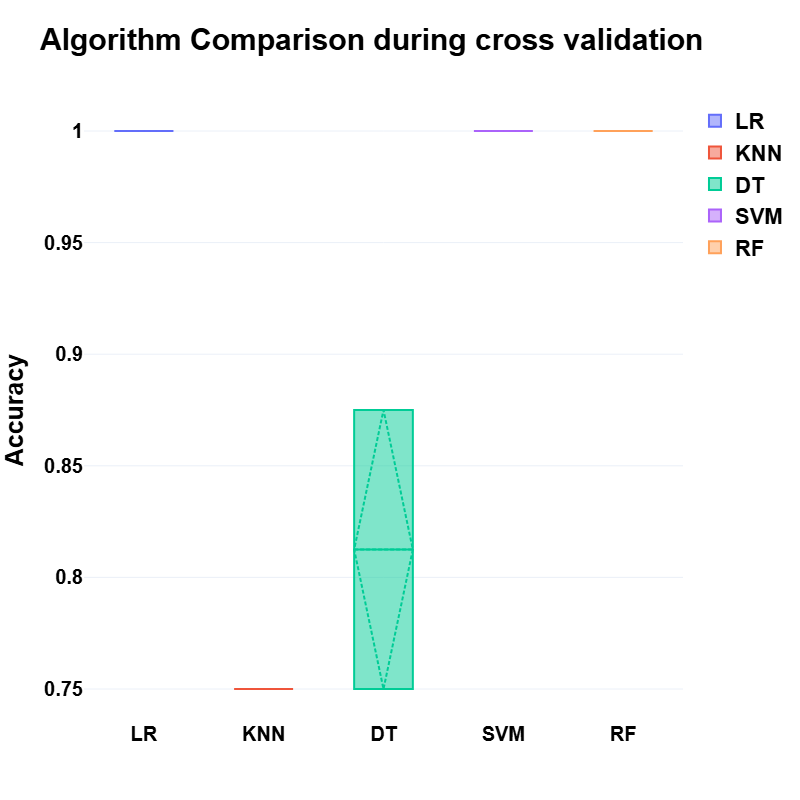}
            \caption[C]%
            {{\small Boxplot chart depicting accuracies obtained by machine learning models during cross-validation}} 
            \label{fig:bx_bc}
        \end{subfigure}
        \hfill
        \begin{subfigure}[b]{0.475\textwidth}   
            \centering 
            \includegraphics[width=0.7\textwidth]{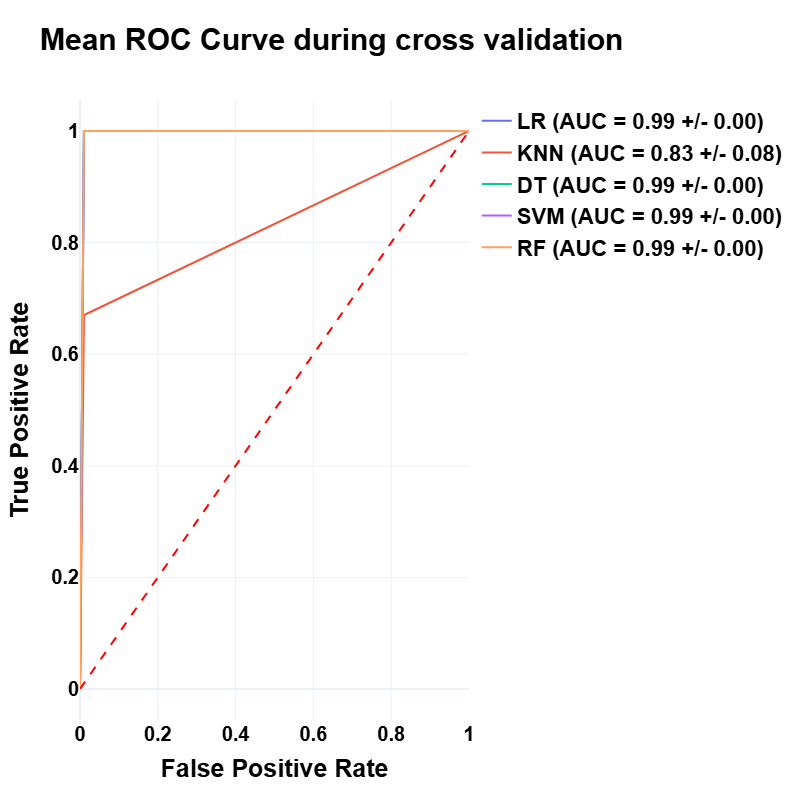}
            \caption[D]%
            {{\small ROC curves obtained by machine learning models}}    
            \label{fig:roc_bc}
        \end{subfigure}
        \caption[]
        {Results of E-ABIN: Subfigure (a) reports the visualization of 2-dimensional gene expression embeddings obtained using PCA; (b) reports the gene expression using TSNE - each point represents a sample, while colours indicate the class.  Subfigure (c) reports the accuracy of the differents models as boxplots. Finally subfigure  (d) depicts the  ROC curves obtained during machine learning models cross-validation.} 
        \label{fig:analysis_bc}
\end{figure*}

E-ABIN takes into input the GSE37817 GEO series matrix. The DNA Methylation values associated with the patient diagnostic class are reported in this matrix. Genes were mapped into known gene names to enhance interpretability. The E-ABIN preprocessing phase handles missing gene values using a class-dependent mode imputation strategy to preserve inherent differences between normal and abnormal methylation.
To visualize each patient methylation profile in two dimensions, dimensionality reduction techniques such as T-distributed Stochastic Neighbor Embedding (t-SNE) \cite{tsne} and Principal Component Analysis (PCA) \cite{pca} were used. An example of a 2D representation of the bladder cancer dataset is depicted in Fig. \ref{fig:pca_bc} and \ref{fig:tsne_bc}, for the PCA and TSNE, respectively. Points were colored based on patient diagnosis to highlight the good separability of this dataset.

Because both embedding functions  showed a remarkable separability between methylation profiles of the two classes, machine learning models might perform well in the binary classification task. 
Given the small dataset size, obtaining good performance using GNNs is not trivial, and hyperparameter optimization might be required. We fixed the same set of hyperparameters for the GAAN and GAE models, while leaving GCN with their own hyperparameter sets. Table \ref{tab:bc_hp} lists the values of the hyperparameters used in this experiment.

\begin{table}[!h]
    \centering
    \footnotesize
    \begin{tabular}{|c|c|c|c|c|c|c|c|c|c|c|}
    \hline
    Model & Epoch & Learning rate & Layers & Noise dim. & Hidden dim. & Dropout & Activation & Backbone & Contamination & Batch \\ \hline
    GAE & 200 & 0.00005 & 2 & None & 128 & 0.3 & ReLU & GCN & 0.375 & 1 \\ \hline
    GAAN & 200 & 0.00005 & 2 & 64 & 128 & 0.3 & ReLU & GCN & 0.375 & 1 \\ \hline
    GCN & 2000 & 0.0005 & 1 & None & 128 & 0.0 & ReLU & GCNConv & None  & 1 \\ \hline
    \end{tabular}
    \caption{Hyperparameters used for bladder cancer detection experiments.}
    \label{tab:bc_hp}
\end{table}

To train both machine learning and deep learning models, the entire dataset was split into a training-test set using a randomized train-test split of 70-30\%. During the machine learning analysis, a 5-fold stratified cross-validation was applied to the training set to assess the overall performance of each model. 

Binary node and graph classifications were performed, and performances were compared against machine-learning ones, as reported in Table \ref{tab:bc_results}. 

Fig. \ref{fig:analysis_bc} shows some of the graphical results obtained using the E-ABIN. In particular:
\begin{itemize}
    \item Fig. \ref{fig:pca_bc} shows a visualization of two-dimensional PCA embeddings of the input methylation dataset. Purple points represent healthy control patients, while orange points indicate anomalous ones. Control patients clustered relatively closely together in the upper left quadrant, suggesting similar gene expression patterns. In contrast, pathological patients were widely scattered across the plot, with many appearing far from the healthy cluster. This clear separation indicates that PCA effectively captured the differences between the normal and anomalous patterns in the bladder cancer dataset.
    \item Fig. \ref{fig:tsne_bc} reports the visualization two-dimensional TSNE embeddings of the input dataset. Compared to PCA, t-SNE has created a more tight cluster for anomalous patients (orange). While healthy control patients (purple) are not clustered tightly, with two samples located far away from the cluster. This suggests that there may be different subtypes of healthy methylation profiles in the dataset.
    \item Fig. \ref{fig:bx_bc} reports accuracy values obtained by each machine learning model during cross-validation. The Decision Tree, Random Forrest and Support Vector Machine models shows better performance accuracy during cross-validation while KNN show lower accuracy values.
    \item Fig. \ref{fig:roc_bc} reports ROC curves and the Area Under the Curve (AUC) obtained by each machine learning model during cross-validation. Since our dataset is highly unbalanced (few control patients compared to the anomalous ones), the AUC metric was computed to assess model performances. The results obtained indicate that LR, SVM, and RF models achieve nearly perfect classification with AUC values of 0.99. The DT model shows the poorest performance with the lowest AUC of 0.71 and the highest standard deviation (±0.29), indicating inconsistent performance across validation folds.
\end{itemize}

\begin{figure*}[!h]
        \centering
        \begin{subfigure}[b]{0.6\textwidth}
            \centering
            \includegraphics[width=0.8\textwidth, keepaspectratio]{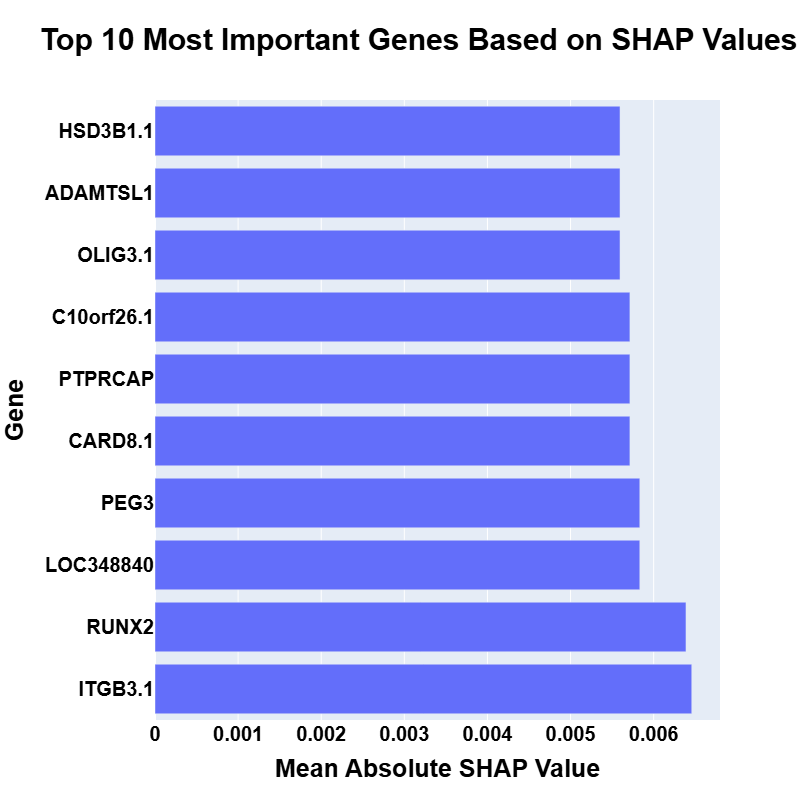}
            \caption[A]%
            {{\small Shap feature importance for bladder cancer classification using Random Forest}}    
            \label{fig:subfig_a}
        \end{subfigure}
        \hfill
        \begin{subfigure}[b]{0.38\textwidth}  
        \centering 
        \begin{subfigure}{0.8\textwidth}
            \centering
            \includegraphics[width=0.9\textwidth, keepaspectratio]{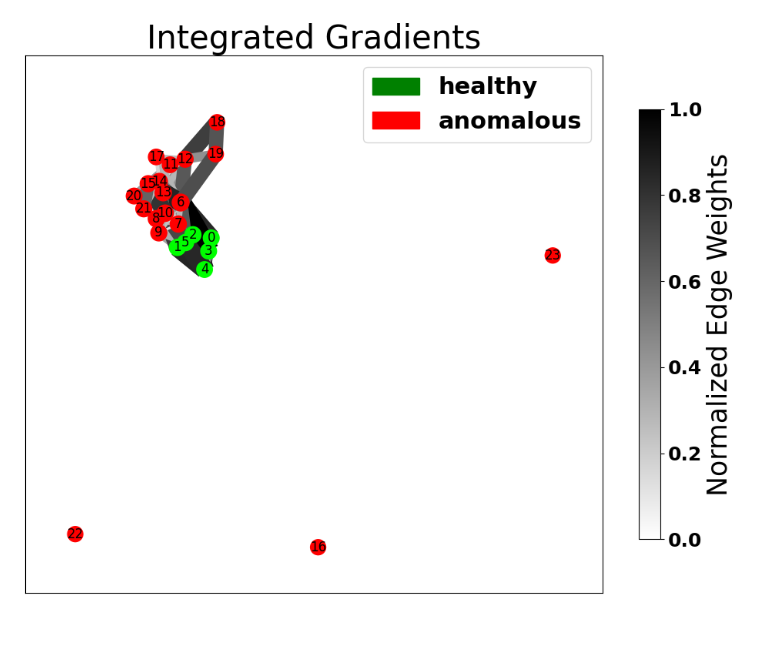}
        \end{subfigure}
        \begin{subfigure}{0.8\textwidth}
            \centering
            \includegraphics[width=0.9\textwidth, keepaspectratio]{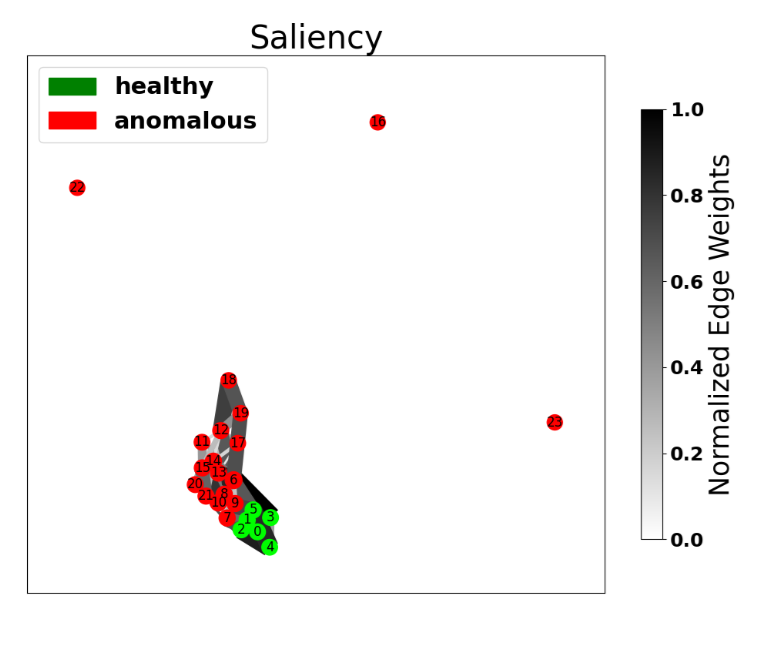}
        \end{subfigure}
        
        \caption[B]{{\small Integrated gradients and saliency maps explanations of GCN model.}}    
        \label{fig:subfig_b}
        \end{subfigure}
        \begin{subfigure}[b]{\textwidth}   
            \centering 
            \includegraphics[width=\textwidth, keepaspectratio]{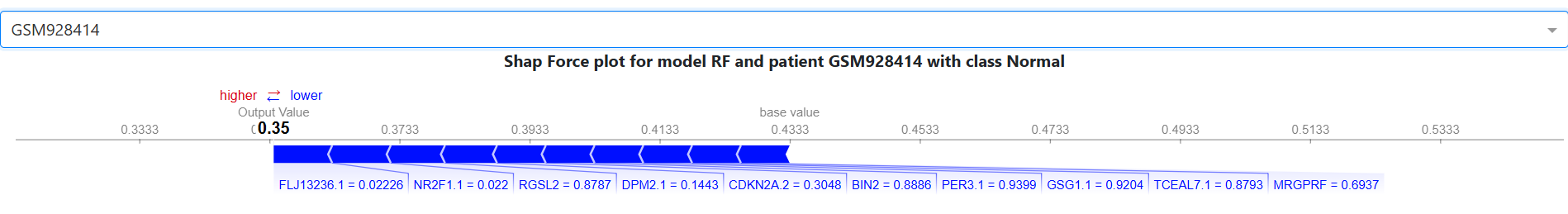}
            \caption[C]%
            {{\small Shap force plot obtained to explain the Random Forest prediction of a Healthy (Normal) patient GSM928414.}}    
            \label{fig:subfig_c}
        \end{subfigure}
        
        \begin{subfigure}[b]{\textwidth}   
            \centering 
            \includegraphics[width=\textwidth, keepaspectratio]{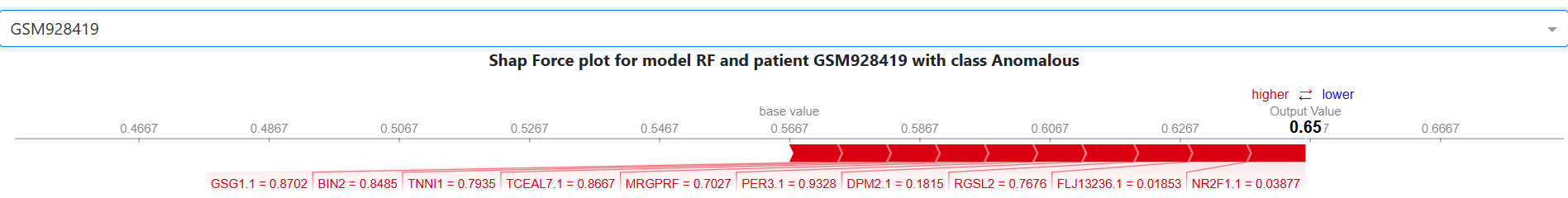}
            \caption[D]%
            {{\small Shap force plot obtained to explain the Random Forest prediction of a Abnormal (bladder cancer) patient GSM928419.}}    
            \label{fig:subfig_d}
        \end{subfigure}
        \hfill
        \caption[Explainability results]
        {\small Explainability figures obtained using E-ABIN. (a) The top 10 most important genes for identifying the anomalous class in the studied dataset. This plot can be obtained using ML models with SHAP explainer or GNN models with GNNExplainer. (b) Integrated gradients and Saliency maps were obtained by using Captum to explain the GCN model. (c) and (d) depicts SHAP force plot obtained for the interpretation of a "healthy" and an "abnormal" prediction, respectively,  made by a Random Forest ML model.} 
        \label{fig:exp_bc}
\end{figure*}

Interpretations of model predictions are available for both machine-learning and deep-learning models. Fig. \ref{fig:exp_bc} shows the options available for this explanation. Fig. \ref{fig:subfig_a} depicts the first ten most important gene expression abnormalities caused by DNA Methylation obtained by applying the SHAP feature importance to machine learning models. From the obtained list of key genes, 5 had an already known correlation with bladder cancer.

Fig. \ref{fig:subfig_c} and. \ref{fig:subfig_d} depict SHAP force plots obtained from the machine learning models to predict a single Healthy and Abnormal patient respectively. The same feature importance can be obtained using GNNExplainer for the GNN models. Finally, important edges were highlighted in Fig. \ref{fig:subfig_b} using the integrated gradients and saliency maps methods.

One limitation of the current deep learning workflow is the absence of a module that automatically optimizes hyperparameters.  Another limitation is the lack of full reproducibility of deep learning performance results due to the random nature of the GAAN model, also confirmed by the authors of the torch and PyGOD library, and different dataset splitting across different runs. To overcome this limitation, E-ABIN is provided with a deterministic modality that enables the reproducibility of our results.

\begin{table}[h]
    \centering
    \footnotesize
    \begin{tabular}{|l|c|c|c|c|c|r|}
    \hline
    Model & Acc \% & f1 & Sens. \%  & Spec. \%  & AUC & Prec \%  \\ \hline
    LR & 100 & 1.00 & 100 & 100 & \textbf{1.00} & 100 \\ \hline
    SVM & 100 & 1.00 & 100 & 100 & \textbf{1.00} & 100 \\ \hline
    RF & 100 & 1.00 & 100 & 100 & \textbf{1.00} & 100 \\ \hline
    DT & 100 & 1.0 & 100 & 100 & \textbf{1.0} & 100 \\ \hline
    KNN & 75 & 0.86 & 100 & 0 & 0.5 & 75 \\ \hline
    GAAN (node) & 50 & 0.44 & 28.57 & 100 & 0.64 & 100 \\ \hline
    GAE (node) & 50 & 0.44 & 28.57 & 100 & 0.64 & 100 \\ \hline
    GCN (node) & 100 & 1.0 & 100 & 100 & \textbf{1.0} & 100 \\ \hline
    GAAN (ISNs) & 75 & 0.8 & 66.67 & 100 & \textit{0.83} & 100 \\ \hline
    \end{tabular}
    \caption{Comparison of E-ABIN algorithms in bladder cancer patient classification task. Best and second best model by ROC AUC score are highlighted in bold and underline, respectively.}
    \label{tab:bc_results}
\end{table}

\section{Discussion}
\label{sec:discussion}

\textbf{E-ABIN} demonstrated strong performance in the analysis of both gene expression and DNA methylation datasets. The framework accurately classified individuals as either healthy or exhibiting abnormalities in their molecular profiles. Crucially, through its integrated explainability module, E-ABIN enables clinicians and researchers to interpret these predictions in biologically meaningful ways, offering multiple layers of insight. At the patient level, the model identifies the most influential genes contributing to each prediction, allowing clinicians to evaluate which molecular features may be causally implicated in disease onset or progression. At the disease level, E-ABIN provides aggregated insights across patient cohorts, highlighting genes that consistently contribute to distinguishing pathological from non-pathological profiles. These results support the identification of potential diagnostic and prognostic markers.

To illustrate the system's capabilities, we analyzed a DNA methylation dataset containing samples from healthy individuals and patients with bladder cancer. Both machine learning and deep learning models embedded in E-ABIN were trained to perform anomaly detection, achieving perfect predictive accuracy on the test set. Even non-trainable dimensionality reduction techniques such as PCA and t-SNE revealed clear separability between the two groups, underscoring the biological relevance of the dataset. In addition to prediction, E-ABIN provided interpretable explanations for each classification, enhancing the transparency and utility of the results in a biomedical setting.

Focusing specifically on the explainability results in bladder cancer, among the ten most influential genes identified by SHAP, seven are supported in the literature as relevant to cancer progression or metastasis. Of these, three genes—\textit{RUNX2}, \textit{ITGB3}, and \textit{CARD8}—have been directly associated with bladder cancer. RUNX2 has been extensively validated and is known to enhance tumor progression via the regulation of glutamine metabolism. Its high expression correlates with poor clinical outcomes and advanced stages, and it plays a role in epithelial-mesenchymal transition and interactions with cancer-associated fibroblasts. ITGB3 has demonstrated clinical utility in prognostic models and is involved in modulating immune infiltration patterns within the tumor microenvironment. CARD8 has been included in bladder cancer prognostic signatures related to immune regulation and contributes to inflammatory pathways through the inflammasome mechanism.

Five additional genes, including \textit{HSD3B1}, \textit{ADAMTSL1}, \textit{PEG3}, \textit{PTPRCAP}, and \textit{PTPRCAP}, have been implicated in other malignancies or metastatic processes. For example, HSD3B1 has been widely studied in prostate cancer and shows enhanced expression in bladder tissues, suggesting hormonal influences that may explain sex-related incidence differences. ADAMTSL1 has been associated with survival outcomes across multiple cancer types and shows methylation changes in other tumors such as breast cancer and rhabdomyosarcoma. PEG3 is a well-characterized tumor suppressor frequently silenced through hypermethylation, with confirmed relevance in bladder, breast, and gynecologic cancers. PTPRCAP has been linked to gastric cancer susceptibility and plays a role in oncogenic signaling pathways involving Src family kinases.

Three genes—\textit{OLIG3}, \textit{C10orf26}, and \textit{LOC348840}—were identified by E-ABIN as potentially relevant, despite limited or no existing experimental validation in bladder cancer. OLIG3, though not directly linked to bladder cancer, belongs to a gene family with known roles in gliomas and neural development. C10orf26 remains poorly characterized but exhibited high SHAP importance, suggesting possible regulatory functions that merit further study. LOC348840, an unannotated locus, also ranked highly in model interpretability analyses and may represent an underexplored genomic element with prognostic potential, in line with prior findings where such loci were later shown to contain cancer-relevant regulatory sequences \cite{Melton2015}.

\section{Conclusion}
\label{sec:results}
\textbf{E-ABIN} integrates machine learning and deep learning methodologies within a unified framework that addresses key limitations of current bioinformatics tools. The system demonstrates robust performance across a range of biological datasets, effectively detecting anomalies in both gene expression and DNA methylation profiles. Crucially, E-ABIN complements these analytical capabilities with an integrated explainability module, allowing for biologically grounded interpretations of the results.

Experimental validation using real-world datasets confirmed E-ABIN’s versatility and effectiveness across different data modalities and scales. In the case of the bladder cancer DNA methylation dataset, classical machine learning models—including logistic regression, support vector machines, and random forests—achieved perfect classification performance, with area under the curve (AUC) values of 1.00. Deep learning models, such as graph convolutional networks (GCNs), also delivered excellent predictive accuracy, underscoring the tool’s adaptability and efficacy.

A distinguishing innovation of E-ABIN lies in its integration of Individual-Specific Networks (ISNs) and convergence/divergence network paradigms, which enable both node-level and graph-level anomaly detection. The framework combines state-of-the-art algorithms—including Graph Adversarial Attributed Networks (GAAN), Graph AutoEncoders (GAEs), and traditional machine learning approaches—with advanced explainability techniques, such as SHAP, GNNExplainer, and Captum. This architecture is presented through a user-friendly graphical interface, lowering technical barriers and making advanced graph-based anomaly detection accessible to a broad range of researchers, including those without extensive computational expertise.

The explainability module provides actionable insights by pinpointing the key genes that most influence disease classification, as demonstrated in the identification of bladder cancer–associated molecular markers. This interpretive capacity bridges the gap between computational analytics and biological understanding, empowering users not only to identify anomalous patient profiles but also to determine the underlying molecular features contributing to those anomalies.

By offering high-performance predictive models alongside interpretable results, E-ABIN establishes a new benchmark for explainable anomaly detection in biological networks. Its ability to combine cutting-edge analytics with transparency supports critical applications in personalized medicine, disease mechanism discovery, and biomarker identification, offering clinicians and researchers a comprehensive, interpretable, and practical tool for molecular network analysis.

\bibliographystyle{plain}
\bibliography{biblio}

\begin{thebibliography}{10}

\bibitem{Akoglu2014GraphBA}
Leman Akoglu, Hanghang Tong, and Danai Koutra.
\newblock Graph based anomaly detection and description: a survey.
\newblock {\em Data Mining and Knowledge Discovery}, 29:626 -- 688, 2014.

\bibitem{anguita2020explainable}
Augusto Anguita-Ruiz, Alberto Segura-Delgado, Rafael Alcal{\'a}, Concepci{\'o}n~M Aguilera, and Jes{\'u}s Alcal{\'a}-Fdez.
\newblock explainable artificial intelligence (xai) for the identification of biologically relevant gene expression patterns in longitudinal human studies, insights from obesity research.
\newblock {\em PLoS computational biology}, 16(4):e1007792, 2020.

\bibitem{asnicar2024machine}
Francesco Asnicar, Andrew~Maltez Thomas, Andrea Passerini, Levi Waldron, and Nicola Segata.
\newblock Machine learning for microbiologists.
\newblock {\em Nature Reviews Microbiology}, 22(4):191--205, 2024.

\bibitem{BANSAL2022724}
Monika Bansal and Dolly Sharma.
\newblock Density-based structural embedding for anomaly detection in dynamic networks.
\newblock {\em Neurocomputing}, 500:724--740, 2022.

\bibitem{bourgeais2021deep}
Victoria Bourgeais, Farida Zehraoui, Mohamed Ben~Hamdoune, and Blaise Hanczar.
\newblock Deep gonet: self-explainable deep neural network based on gene ontology for phenotype prediction from gene expression data.
\newblock {\em BMC bioinformatics}, 22:1--25, 2021.

\bibitem{brooks2024challenges}
Thomas~G Brooks, Nicholas~F Lahens, Antonijo Mr{\v{c}}ela, and Gregory~R Grant.
\newblock Challenges and best practices in omics benchmarking.
\newblock {\em Nature Reviews Genetics}, 25(5):326--339, 2024.

\bibitem{Clough2016geo}
Emily Clough and Tanya Barrett.
\newblock {\em The Gene Expression Omnibus Database}, pages 93--110.
\newblock Springer New York, New York, NY, 2016.

\bibitem{ding2022explainability}
Weiping Ding, Mohamed Abdel-Basset, Hossam Hawash, and Ahmed~M Ali.
\newblock Explainability of artificial intelligence methods, applications and challenges: A comprehensive survey.
\newblock {\em Information Sciences}, 615:238--292, 2022.

\bibitem{pca}
Karl~Pearson F.R.S.
\newblock Liii. on lines and planes of closest fit to systems of points in space.
\newblock {\em The London, Edinburgh, and Dublin Philosophical Magazine and Journal of Science}, 2(11):559--572, 1901.

\bibitem{gae}
Thomas~N. Kipf and Max Welling.
\newblock Variational graph auto-encoders, 2016.

\bibitem{gcn}
Thomas~N. Kipf and Max Welling.
\newblock Semi-supervised classification with graph convolutional networks.
\newblock In {\em 5th International Conference on Learning Representations, {ICLR} 2017, Toulon, France, April 24-26, 2017, Conference Track Proceedings}. OpenReview.net, 2017.

\bibitem{kokhlikyan2020captum}
Narine Kokhlikyan, Vivek Miglani, Miguel Martin, Edward Wang, Bilal Alsallakh, Jonathan Reynolds, Alexander Melnikov, Natalia Kliushkina, Carlos Araya, Siqi Yan, and Orion Reblitz-Richardson.
\newblock Captum: A unified and generic model interpretability library for pytorch, 2020.

\bibitem{kuehn2008using}
Heidi Kuehn, Arthur Liberzon, Michael Reich, and Jill~P Mesirov.
\newblock Using genepattern for gene expression analysis.
\newblock {\em Current protocols in bioinformatics}, 22(1):7--12, 2008.

\bibitem{KUIJJER2019226}
Marieke~Lydia Kuijjer, Matthew~George Tung, GuoCheng Yuan, John Quackenbush, and Kimberly Glass.
\newblock Estimating sample-specific regulatory networks.
\newblock {\em iScience}, 14:226--240, 2019.

\bibitem{pygod}
Kay Liu, Yingtong Dou, Xueying Ding, Xiyang Hu, Ruitong Zhang, Hao Peng, Lichao Sun, and Philip~S. Yu.
\newblock {PyGOD}: A {Python} library for graph outlier detection.
\newblock {\em Journal of Machine Learning Research}, 25(141):1--9, 2024.

\bibitem{liu2022bond}
Kay Liu, Yingtong Dou, Yue Zhao, Xueying Ding, Xiyang Hu, Ruitong Zhang, Kaize Ding, Canyu Chen, Hao Peng, Kai Shu, Lichao Sun, Jundong Li, George~H. Chen, Zhihao Jia, and Philip~S. Yu.
\newblock {BOND}: Benchmarking unsupervised outlier node detection on static attributed graphs.
\newblock In {\em Thirty-sixth Conference on Neural Information Processing Systems Datasets and Benchmarks Track}, 2022.

\bibitem{pygodbench}
Kay Liu, Yingtong Dou, Yue Zhao, Xueying Ding, Xiyang Hu, Ruitong Zhang, Kaize Ding, Canyu Chen, Hao Peng, Kai Shu, Lichao Sun, Jundong Li, George~H Chen, Zhihao Jia, and Philip~S Yu.
\newblock {BOND}: Benchmarking unsupervised outlier node detection on static attributed graphs.
\newblock In S.~Koyejo, S.~Mohamed, A.~Agarwal, D.~Belgrave, K.~Cho, and A.~Oh, editors, {\em Advances in Neural Information Processing Systems}, volume~35, pages 27021--27035. Curran Associates, Inc., 2022.

\bibitem{shap}
Scott~M Lundberg and Su-In Lee.
\newblock A unified approach to interpreting model predictions.
\newblock In I.~Guyon, U.~V. Luxburg, S.~Bengio, H.~Wallach, R.~Fergus, S.~Vishwanathan, and R.~Garnett, editors, {\em Advances in Neural Information Processing Systems 30}, pages 4765--4774. Curran Associates, Inc., 2017.

\bibitem{Xiaoxiao2023gad}
Xiaoxiao Ma, Jia Wu, Shan Xue, Jian Yang, Chuan Zhou, Quan~Z. Sheng, Hui Xiong, and Leman Akoglu.
\newblock A comprehensive survey on graph anomaly detection with deep learning.
\newblock {\em IEEE Transactions on Knowledge and Data Engineering}, 35(12):12012--12038, 2023.

\bibitem{Melton2015}
Collin Melton, Jason Reuter, Damek Spacek, and Michael Snyder.
\newblock Recurrent somatic mutations in regulatory regions of human cancer genomes.
\newblock {\em Nature genetics}, 47, 06 2015.

\bibitem{morabito2023genes}
Fortunato Morabito, Carlo Adornetto, Paola Monti, Adriana Amaro, Francesco Reggiani, Monica Colombo, Yissel Rodriguez-Aldana, Giovanni Tripepi, Graziella D’Arrigo, Claudia Vener, et~al.
\newblock Genes selection using deep learning and explainable artificial intelligence for chronic lymphocytic leukemia predicting the need and time to therapy.
\newblock {\em Frontiers in Oncology}, 13:1198992, 2023.

\bibitem{noto2015csax}
Keith Noto, Saeed Majidi, Andrea~G Edlow, Heather~C Wick, Diana~W Bianchi, and Donna~K Slonim.
\newblock Csax: Characterizing systematic anomalies in expression data.
\newblock {\em Journal of Computational Biology}, 22(5):402--413, 2015.

\bibitem{rambaud2023binaryclass}
Philippe Rambaud, Adel Taleb, Raphael Fauches, Arpad Rimmel, Joanna Tomasik, and Jean Bergounioux.
\newblock { Binary Classification vs. Anomaly Detection on Imbalanced Tabular Medical Datasets }.
\newblock In {\em 2023 Congress in Computer Science, Computer Engineering, \& Applied Computing (CSCE)}, pages 1--5, Los Alamitos, CA, USA, July 2023. IEEE Computer Society.

\bibitem{rosati2024differential}
Diletta Rosati, Maria Palmieri, Giulia Brunelli, Andrea Morrione, Francesco Iannelli, Elisa Frullanti, and Antonio Giordano.
\newblock Differential gene expression analysis pipelines and bioinformatic tools for the identification of specific biomarkers: A review.
\newblock {\em Computational and structural biotechnology journal}, 2024.

\bibitem{rousseeuw2018statistical}
Peter~J. Rousseeuw and Mia Hubert.
\newblock Anomaly detection by robust statistics.
\newblock {\em WIREs Data Mining and Knowledge Discovery}, 8(2):e1236, 2018.

\bibitem{toussaint2024explainable}
Philipp~A Toussaint, Florian Leiser, Scott Thiebes, Matthias Schlesner, Benedikt Brors, and Ali Sunyaev.
\newblock Explainable artificial intelligence for omics data: a systematic mapping study.
\newblock {\em Briefings in Bioinformatics}, 25(1):bbad453, 2024.

\bibitem{tsne}
Laurens van~der Maaten and Geoffrey Hinton.
\newblock Visualizing data using t-sne.
\newblock {\em Journal of Machine Learning Research}, 9(86):2579--2605, 2008.

\bibitem{withnell2021xomivae}
Eloise Withnell, Xiaoyu Zhang, Kai Sun, and Yike Guo.
\newblock Xomivae: an interpretable deep learning model for cancer classification using high-dimensional omics data.
\newblock {\em Briefings in bioinformatics}, 22(6):bbab315, 2021.

\bibitem{zamanzadeh2024deep}
Zahra Zamanzadeh~Darban, Geoffrey~I Webb, Shirui Pan, Charu Aggarwal, and Mahsa Salehi.
\newblock Deep learning for time series anomaly detection: A survey.
\newblock {\em ACM Computing Surveys}, 57(1):1--42, 2024.

\bibitem{zanin2018}
Massimiliano Zanin, Juan~Manuel Tuñas, and Ernestina Menasalvas.
\newblock Understanding diseases as increased heterogeneity: a complex network computational framework.
\newblock {\em Journal of The Royal Society Interface}, 15(145):20180405, 2018.

\end{thebibliography}

\section*{Software Availability}
\label{sec:availability}
Software and User Guide are available at \url{https://github.com/UgoLomoio/E-ABIN}

\newcommand{\appendix}{\setcounter{section}{0}\renewcommand{\thesection}{Appendix \Alph{section}}}
\appendix
\setcounter{section}{0}
\renewcommand{\thesection}{Appendix \Alph{section}}

\begin{biography}{{\color{black!20}\rule{77pt}{77pt}}}{\author{Author Name.} This is sample author biography text. The values provided in the optional argument are meant for sample purposes. There is no need to include the width and height of an image in the optional argument for live articles. This is sample author biography text this is sample author biography text this is sample author biography text this is sample author biography text this is sample author biography text this is sample author biography text this is sample author biography text this is sample author biography text.}
\end{biography}


\end{document}